\pdfoutput=1
\documentclass[11pt]{article}

\usepackage[]{EACL2023}
\usepackage{amsfonts,amssymb}
\usepackage{times}
\usepackage{latexsym}
\usepackage{multirow}
\usepackage[T1]{fontenc}
\usepackage{booktabs}
\usepackage{amsmath}
\usepackage{hyperref}
\usepackage{MnSymbol}

\usepackage[utf8]{inputenc}
\usepackage{subfigure}
\usepackage{parskip}
\usepackage{graphicx}
\usepackage{float}
\usepackage{caption}
\usepackage{diagbox}
\usepackage{makecell}
\usepackage{bbding} 
\usepackage{microtype}

\title{Fine-Tuning Deteriorates General Textual Out-of-Distribution Detection by Distorting Task-Agnostic Features}

\author{Sishuo Chen\textsuperscript{1}, Wenkai Yang\textsuperscript{1},  Xiaohan Bi\textsuperscript{1}, Xu Sun\textsuperscript{2} \\
  \textsuperscript{1}Center for Data Science, Peking University\\
  \textsuperscript{2}MOE Key Laboratory of Computational Linguistics, School of Computer Science, \\ Peking University\\
    \texttt{\{chensishuo,xusun\}@pku.edu.cn} \\
  \texttt{\{wkyang,bxh\}@stu.pku.edu.cn} }

\begin{document}
\maketitle

\begin{abstract}

Detecting out-of-distribution (OOD) inputs is crucial for the safe deployment of natural language processing (NLP) models. 
Though existing methods, especially those based on the statistics in the feature space of fine-tuned pre-trained language models (PLMs), are claimed to be effective, their effectiveness on different types of distribution shifts remains underexplored. 
In this work, we take the first step to comprehensively evaluate the mainstream textual OOD detection methods for detecting semantic and non-semantic shifts. 
We find that: (1) no existing method behaves well in both settings; (2) fine-tuning PLMs on in-distribution data benefits detecting semantic shifts but severely deteriorates detecting non-semantic shifts, which can be attributed to the distortion of task-agnostic features. 
To alleviate the issue, we present a simple yet effective general OOD score named GNOME that integrates the confidence scores derived from the task-agnostic and task-specific representations. 
Experiments show that GNOME works well in both semantic and non-semantic shift scenarios, and further brings significant improvement on two cross-task benchmarks where both kinds of shifts
simultaneously take place. 
Our code is available at \url{https://github.com/lancopku/GNOME}.
\end{abstract}

\section{Introduction} \label{sec:intro}

The pre-training and fine-tuning paradigm based on Transformers \citep{DBLP:conf/nips/VaswaniSPUJGKP17} has achieved tremendous success in various natural language understanding (NLU) tasks \citep{devlin-etal-2019-bert,roberta,qiu2020pre}. 
However, fine-tuned pre-trained language models (PLMs) notoriously suffer from over-confident predictions on out-of-distribution (OOD) inputs \cite{hendrycks2020pretrained}. 
As this issue threats the reliability of NLP models deployed in the open world, textual OOD detection has attracted great attention recently~\citep[etc.]{podolskiy2021revisiting,contrastive_nlp_ood,zhou-etal-2022-knn,duan2022background}, which aims to enable the model to abstain from making unreasonable predictions on OOD data and resort to human intervention.

\begin{figure}[t]
\centering
\includegraphics[width=0.45\textwidth]{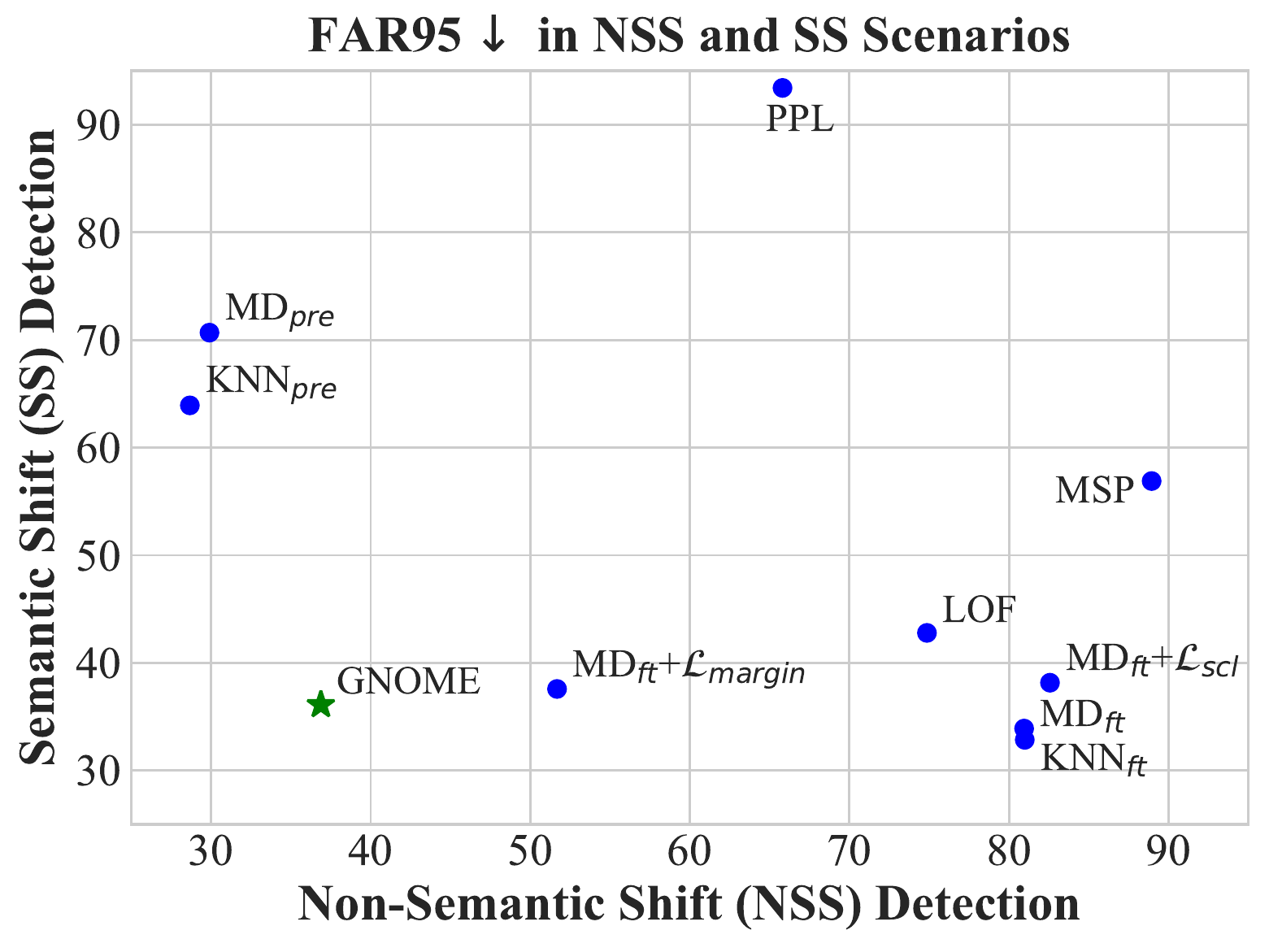}
\caption{OOD detection performance (FAR95$\downarrow$, lower is better) in non-semantic and semantic shift scenarios. No single existing method works well in both scenarios, but our proposed GNOME mitigates the trade-off.}
\label{fig0}
\end{figure}        

Nonetheless, almost all of the current approaches are assessed under certain assumptions about the type of OOD texts. 
One line of works creates in-distribution (ID) and OOD pairs from arbitrary datasets for different tasks~\citep{hendrycks2020pretrained}, while another line assumes that OOD data belong to classes in the ID task but unseen during training, e.g., in intent recognition~\citep{podolskiy2021revisiting}. 
\citet{types} reveal the inconsistency among the evaluation protocols and category the distribution shifts to non-semantic shifts (NSS) and semantic shifts (SS), but a thorough comparison of existing methods in different settings is missing as later works either focus on either detecting NSS~\citep{duan2022background} or SS~\citep{zhou-etal-2022-knn}.

In this work, we systematically evaluate the mainstream textual OOD detection methods on a comprehensive suite of benchmarks covering both  NSS and SS scenarios. 
As shown in Figure~\ref{fig0}, no single method wins across the board.
Notably, the detectors based on the pre-trained features, e.g., the Mahalanobis distance detector MD$_{\text{pre}}$~\citep{xu2021unsupervised}, excel at detecting non-semantic shifts but fail in detecting semantic shifts. 
In contrast, when the PLM is fine-tuned on annotated ID data, the detectors based on fine-tuned features, e.g., MD$_{\text{ft}}$~\citep{podolskiy2021revisiting},  perform well in the SS scenario but disastrously fail in the NSS setting.
These observations uncover an intriguing trade-off: \textit{fine-tuning contributes to the detection of semantic shifts but impairs the detection of non-semantic shifts}.
This trade-off raises two critical research questions:


\textbf{RQ1:} \emph{Why does fine-tuning undermine the detection of non-semantic shifts?}
It is relatively easy to attribute the positive effect of fine-tuning in the SS setting to the learned class-discriminative features~\citep{fort2021exploring}, but it remains unknown why fine-tuning plays a negative role in the NSS setting.
We empirically find that the adverse effect comes from the fact that fine-tuning gradually destructs the pre-trained task-agnostic knowledge
about general linguistic properties,  which are useful cues for the detection of non-semantic shifts.

\textbf{RQ2:} \emph{How to develop a general textual OOD detection method?}
Since the type of distribution shifts is unknown in practice, our findings suggest that a practical method able to detect different kinds of OOD texts is yet to be developed.
To this end, we aggregate the distance scores estimated in the feature space of both pre-trained and fine-tuned models to derive a \textbf{G}e\textbf{N}eral textual \textbf{O}OD \textbf{M}easurement scor\textbf{E} (GNOME) capable of detecting both NSS and SS.
On the suite of benchmarks covering both NSS and SS settings, GNOME (the green star in Figure~\ref{fig0}) surpasses the previous SOTA by 8.13 FAR95 points on average; on two cross-task benchmarks where both kinds of shifts happen simultaneously, GNOME reduces the average FAR95 by 4.88 points.
Note that GNOME is not meant to be a SOTA method in all settings but rather a simple, principled way to get  reasonable detection performance under various kinds of distribution shifts—we hope our analysis inspires better approaches for general textual OOD detection.


\section{Related Work} \label{sec:related_work}

OOD detection aims to detect abnormalities coming from a different distribution from the training data so that the model can refuse to make predictions on them~\citep{amodei2016concrete,yang2021oodsurvey}. 
Since it is essential for the security of machine learning models deployed in the open-world environment, OOD detection has gained great attention, first in computer vision (CV). 
We categorize the mainstream OOD detection methods into three groups by way to derive confidence scores:
(1) \textit{confidence-based methods} using the output probabilities of classifiers trained on in-distribution (ID) data \citep{maxprob,de,odin,liu2020Energy};
(2) \textit{density-based methods} using density scores derived from generation models \citep{zong2018deep,ratio,xiao2020likelihood};
(3) \textit{distance-based methods} using the distance statistics in the feature space of neural networks \citep{maha,fssd,sun2022knn}.

Following the progress in CV, textual OOD detection based on PLMs has also attracted increasing attention.
\citet{hendrycks2020pretrained} show that the maximum softmax probability (MSP) score~\citep{maxprob} is a strong baseline for PLMs, followed by a group of works on confidence-based textual OOD detection \citep{li2021kfolden,shen2021enhancing,yilmazd2u}.
As for the density-based branch, \citet{gangal2020likelihood} and \citet{types} apply the idea to textual OOD detection by leveraging language models such as LSTM~\citep{lstm} and GPT-2~\citep{gpt2}.
Regarding the distance-based methods, \citet{podolskiy2021revisiting} revisit the Mahalanobis distance-based detector~\citep{maha} for textual OOD detection based on fine-tuned PLMs and achieve performance gains over confidence-based methods, which is then further improved by introducing contrastive regularization~\citep{contrastive_nlp_ood}, utilizing nearest-neighbor distance~\citep{zhou-etal-2022-knn}, and leveraging intermediate features~\citep{chen2022holistic}.

Nonetheless, the NLP community lacks uniform evaluation criteria for OOD detection.
Generally, ID/OOD pairs for evaluation are constructed in three ways:
(1) \textit{the non-semantic shift (NSS) setting} (a.k.a., the background shift setting) \citep{li2021kfolden,types,duan2022background}, where ID and OOD data consist of the same semantic classes but differ in background information,\footnote{Although the model can also make predictions on the samples with only non-semantic shifts, the accuracy tends to significantly drop. As the cost of wrong predictions is great in safety-critical scenarios, a conservative method for handling these samples by rejecting them is practical.} e.g., tweets as ID and Wikipedia comments as OOD in toxicity detection;
(2) \textit{the semantic shift (SS) setting}~\citep{podolskiy2021revisiting,zhou-etal-2022-knn}, where OOD data are composed of unseen classes belonging to the ID task, e.g., new classes in intent classification;
(3) \textit{the cross-task setting} \citep{hendrycks2020pretrained,contrastive_nlp_ood}, where the ID and OOD data are from datasets for different tasks and both semantic and non-semantic shifts happen, e.g., sentiment analysis data as ID and news classification data as OOD.
\citet{types} first notice the inconsistency and compare confidence-based and density-based methods in NSS and SS settings, but they neglect the crucial branch of distance-based methods and the cross-task setting.
In this work, we fill in this gap by presenting a comprehensive evaluation and developing a general textual OOD score motivated by our observations.\looseness=-1



\section{Observations and Explanations}

In this section, we first give preliminaries in \S~\ref{subsec:3.1}.
Then we introduce our benchmark for evaluating textual OOD detection (\S~\ref{subsec:3.2}) and the evaluated methods (\S~\ref{subsec:3.3}).    
Finally, we present the evaluation results (\S~\ref{subsec:3.4}) and our interpretation of the observed trade-off between NSS and SS scenarios (\S~\ref{subsec:3.5}).

\subsection{Preliminaries} \label{subsec:3.1}

\paragraph{Problem Formulation}

The OOD detection problem can be formulated as a binary classification problem to decide whether an input example $\mathbf{x}$ belongs to the training data distribution $\mathcal{P}_{\text {in }}$ (ID) or not (OOD).  
An OOD detector $D$  makes decisions for the input $\mathbf{x}$ based on the following formula:
\begin{equation}
\centering
D(\mathbf{x})= \begin{cases}\text { ID } & \text { if } S(\mathbf{x}) \geq \gamma \\ \text { OOD } & \text { if } S(\mathbf{x})<\gamma\end{cases}, 
\end{equation}
where $S(\mathbf{x})$ is the confidence score output by the detector  and $\gamma$ is the threshold chosen by the user.
\paragraph{Metrics}

We adopt two widely-used metrics AUROC and FAR95 following prior works \citep{podolskiy2021revisiting,contrastive_nlp_ood}.
AUROC can be interpreted as the probability that the model ranks a random ID  sample higher than a random OOD sample, and FAR95 is the proportion of negative samples (OOD) wrongly judged as positive (ID) when the true positive rate is 95\%. 
Higher AUROCs and lower FAR95s indicate better performance.

\paragraph{Notations} Assume $M_{\theta}$ is a PLM where $\theta$ denotes its parameters and $\mathbf{z}=M(\mathbf{x})$ denotes the feature vector for the input sample $\mathbf{x}$  derived from $M$ (e.g., the last-layer CLS embedding in Transformers). 
For a classification task with $C$ classes, the user fine-tunes $M$ together with a classification head $h$ and get the fine-tuned model $F_{\theta^{\ast},h}=h \circ M_{\theta^{\ast}}$ where $\theta^{\ast}$ denotes the fine-tuned parameters.
The output of $F$ is $F_{\theta^{\ast},h}\left(\mathbf{x}\right) = \left(p_1\left(\mathbf{x}\right),p_2\left(\mathbf{x}\right),\ldots,p_C\left(\mathbf{x}\right)\right)^T$, which denotes the predicted probabilities.


\begin{table}[t]
\centering
\resizebox{0.45\textwidth}{!}{
\begin{tabular}{@{}cccc@{}}
\toprule
\textbf{Setting}                                                              & \textbf{Task}                                                                             & \textbf{ID}   & \textbf{OOD}  \\ \midrule
\multirow{4.5}{*}{\begin{tabular}[c]{@{}c@{}}Non-Semantic\\ Shift\end{tabular}} & \multirow{2}{*}{\begin{tabular}[c]{@{}c@{}}Sentiment\\ Analysis\end{tabular}}             & SST-2         & IMDB          \\
                                                                              &                                                                                           & IMDB          & SST-2         \\ \cmidrule(l){2-4}
                                                                              & \multirow{2}{*}{\begin{tabular}[c]{@{}c@{}}Toxic\\ Detection\end{tabular}}                & Twitter       & Jigsaw        \\
                                                                              &                                                                                           & Jigsaw        & Twitter       \\ \midrule
\multirow{4.5}{*}{\begin{tabular}[c]{@{}c@{}}Semantic\\ Shift\end{tabular}}     & \multirow{2}{*}{\begin{tabular}[c]{@{}c@{}}News \\ Categorization\end{tabular}}           & AGNews        & AGNews$_{\text{OOD}}$        \\
                                                                              &                                                                                           & NC & NC$_{\text{OOD}}$ \\  \cmidrule(l){2-4}
                                                                              & \multirow{2}{*}{\begin{tabular}[c]{@{}c@{}}Dialogue Intent\\  Classification\end{tabular}} & ROSTD         & ROSTD$_{\text{OOD}}$         \\
                                                                              &                                                                                           & CLINC         & CLINC$_{\text{OOD}}$         \\ \bottomrule
\end{tabular}}
\caption{The architecture of the constructed suite of benchmarks categorized as either non-semantic shift (NSS) or semantic shift (SS). NC is short for the News Category dataset.}
\label{tab:benchmark_design}
\end{table}

\subsection{Benchmark Construction} \label{subsec:3.2}

We aim to build ID/OOD pairs where either the non-semantic shift (NSS) or the semantic shift (SS) dominates so that we can fairly compare existing methods on the ability to detect these two kinds of shifts separately. 
(1) For NSS, we choose SST-2 \citep{sst2} and IMDB \citep{imdb} for  sentiment analysis, and Twitter \citep{twitter} and Jigsaw\footnote{Available at this \href{https://www.kaggle.com/c/jigsaw-toxic-comment-classification-challenge}{link}.} for toxicity detection. Among the four, any two datasets from the same task can be regarded as an ID/OOD pair.
(2) For SS, we use four datasets: New Category (NC) \citep{misra2018news}, AGNews \citep{agnews}, ROSTD \citep{gangal2020likelihood}, and CLINC \citep{clinc}.
For each dataset, we use some classes as ID and the remaining classes as OOD. 
We show the architecture of the constructed suite of benchmarks categorized as either non-semantic shift (NSS) or semantic shift (SS) in Table~\ref{tab:benchmark_design} and
More details can be found in Appendix~\ref{app:datasets}.
Compared with \citet{types},  we additionally include toxicity detection data for NSS and intent recognition data for SS, which make the suite of benchmarks more representative of real-world scenarios.\looseness=-1

\subsection{Evaluated Baselines} \label{subsec:3.3}

\paragraph{OOD Detection Methods}

We evaluate the mainstream methods as follows.
(1) For \textit{confidence-based methods}, we test the MSP baseline ($S(\mathbf{x})=\max_{y \in \left\{1,2,\ldots,C\right\}}p_y(\mathbf{x})$) \citep{maxprob} and its three variants: Scaling \citep{odin}, Energy Score \citep{liu2020Energy}, and D2U \citep{yilmazd2u};
(2) For \textit{density-based methods}, we evaluate the PPL method \citep{types} using the GPT-2 model for language modeling ($S(\mathbf{x})=1/\operatorname{PPL}(\mathbf{x})$);
(3) For \textit{distance-based methods}, we test the LOF method \citep{lin2019deep} that trains a local outlier detector on fine-tuned features of ID data, and the basic variants of the Mahalanobis detector (MD): MD$_{\text{pre}}$ \citep{xu2021unsupervised} built on pre-trained features 
($\mathbf{z}=M_{\theta}(\mathbf{x})$) and MD$_{\text{ft}}$ \citep{podolskiy2021revisiting} built on fine-tuned features ($\mathbf{z}=M_{\theta^{\ast}} (\mathbf{x})$). Also, we evaluate two variants of MD built on features derived from PLMs fine-tuned with supervised contrastive and margin-based auxiliary targets \citep{contrastive_nlp_ood}, namely $\text{MD}_{\text{ft}} $ + $ \mathcal{L}_{\text{scl}} $ and $\text{MD}_{\text{ft}} $ + $ \mathcal{L}_{\text{margin}} $. Generally, the confidence score in MD is formulated as:
\begin{equation} 
\small
\begin{gathered}
\text{MD}(\mathbf{x})=\min_{c \in \left\{1,2,\ldots,C\right\}}  \left(\mathbf{z}-\mu_{c}\right)^{T} \Sigma^{-1}\left(\mathbf{z}-\mu_{c}\right), \\
 S(\mathbf{x}) = -\text{MD}(\mathbf{x}),
\end{gathered}
\end{equation}
where $\mu_c$ is the class centroid for class $c$ and $\Sigma$ is the global covariance matrix ($\mu$ and $\Sigma$ can be estimated on ID training data).
Besides, we evaluate the nearest-neighbor detectors~\citep{sun2022knn} based on pre-trained ($\text{KNN}_{\text{pre}}$) and fine-tuned features ($\text{KNN}_{\text{ft}}$).
We refer readers to Appendix~\ref{app:details} for more details about the baselines.

\paragraph{Model Configuration}
For the methods based on fine-tuned PLMs, we build text classifiers by fine-tuning the RoBERTa$_{\text{base}}$ \citep{roberta} model (110M parameters) on annotated ID data. 
For MD$_{\text{pre}}$, we use the pre-trained RoBERTa$_{\text{base}}$ model. 
For the PPL method, we fine-tune the GPT-2$_{\text{small}}$ model (117M parameters) for language modeling on ID data.
More details  can be found in Appendix~\ref{app:id_acc}.

\begin{table}[t]{
\centering
\resizebox{0.48\textwidth}{!}{
\begin{tabular}{@{}ll|c|cc@{}}
\toprule
\textbf{Category}                     & \textbf{Method} & \textbf{Avg.} & \textbf{NSS} & \textbf{SS} \\ \midrule
\multirow{4}{*}{Confidence} &  MSP                &   71.63/72.92           & 65.47/88.94 & 77.78/56.89             \\
                                  &   Scaling             &     71.96/71.62        &     65.45/88.94         &         78.47/54.30     \\
                                  & Energy            &        71.75/71.63      &     64.90/89.05         &       78.61/54.20      \\
                                  & D2U                &    71.99/71.49          &       65.47/88.94       &    78.52/54.04         \\ \midrule
Density                    &     PPL            &      67.65/79.61        &   74.28/65.81           &      61.03/93.42       \\ \midrule
\multirow{6}{*}{Distance}   
    &     MD$_{\text{ft}}$            &     80.39/57.42         &        72.25/80.95      &  88.54/33.89            \\
    &     $\text{MD}_{\text{ft}} $ + $ \mathcal{L}_{\text{scl}} $    &   82.22/60.36      &          76.71/82.57    &   87.73/38.15                \\ 
    &      $\text{MD}_{\text{ft}} $ + $ \mathcal{L}_{\text{margin}} $            &        \textbf{86.50}/\textbf{44.63}      &  85.45/51.68             &   87.54/37.57          \\ 
&     MD$_{\text{pre}}$            &      83.76/50.29        &  \textbf{93.29}/29.90             & 74.22/70.68             \\
&     KNN$_{\text{ft}}$            &      81.02/56.91     &     72.58/80.99         &      \textbf{89.47}/\textbf{32.84}       \\
&     KNN$_{\text{pre}}$            &    85.69/46.29       &      92.66/\textbf{28.67}       &       78.72/63.92     \\
\bottomrule
\end{tabular}}
\caption{The performance (AUROC$\uparrow$/FAR95$\downarrow$ values in percentage) of the evaluated approaches. All results are averaged over five random seeds, and best results are highlighted in \textbf{bold}. We report results averaged on the ID/OOD pairs in NSS and SS setting in the last two columns, respectively, and report the results averaged on all eight benchmarks in the third column. See full results on each benchmark in Table~\ref{tab:main_results} and Appendix~\ref{app:results}. }
\label{tab:eval_results_brief}
}
\end{table}

\subsection{Evaluation Results and Findings} \label{subsec:3.4}

We display the main evaluation results in Table~\ref{tab:eval_results_brief}.
As shown, the confidence-based methods underperform the density-based method PPL in the NSS setting, while they outrival PPL in the SS setting, in line with the observations in \citet{types}.
Notably, we notice that the distance-based methods achieve the best results in both NSS and SS settings.
Concretely, MD$_\text{pre}$ and  KNN$_\text{pre}$ built on pre-trained features are the best in the NSS setting, while MD$_\text{ft}$ and KNN$_\text{ft}$ built on fine-tuned features are the best in the SS setting. However, no single method wins across the board.
Thus, we draw an intriguing trade-off: \emph{In textual OOD detection, fine-tuning PLMs on ID data boosts semantic shift detection but impairs non-semantic shift detection.}

\begin{figure}[t]
    \centering
    \begin{subfigure}[Non-Semantic Shift: SST-2 (ID) vs. IMDB (OOD)]{
         \includegraphics[width=0.45\textwidth]{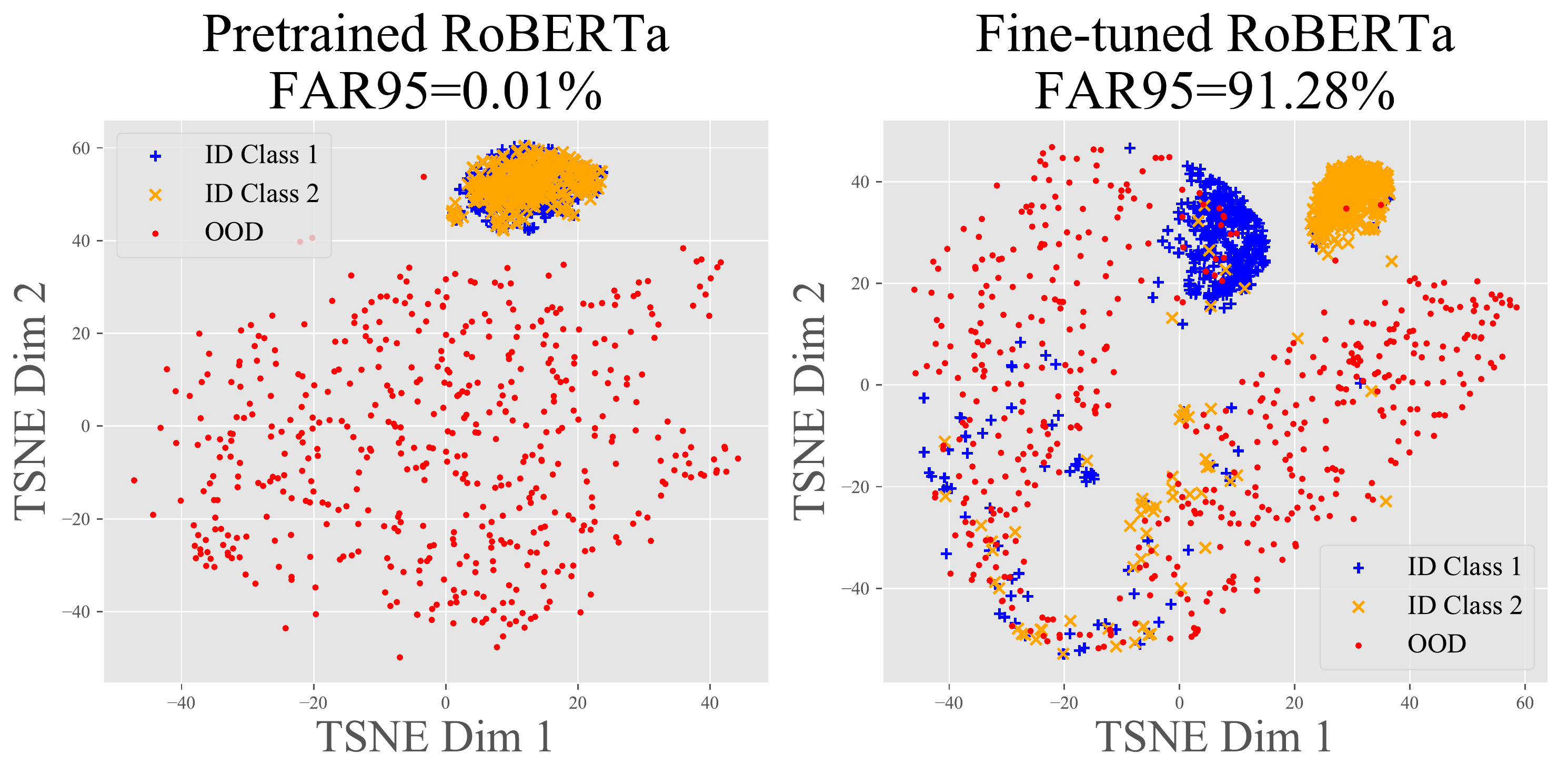}
         \label{fig1a}}
    \end{subfigure}
    
    \begin{subfigure}[Semantic Shift: ROSTD (ID) vs. ROSTD$_{\text{OOD}}$ (OOD)]{
         \includegraphics[width=0.45\textwidth]{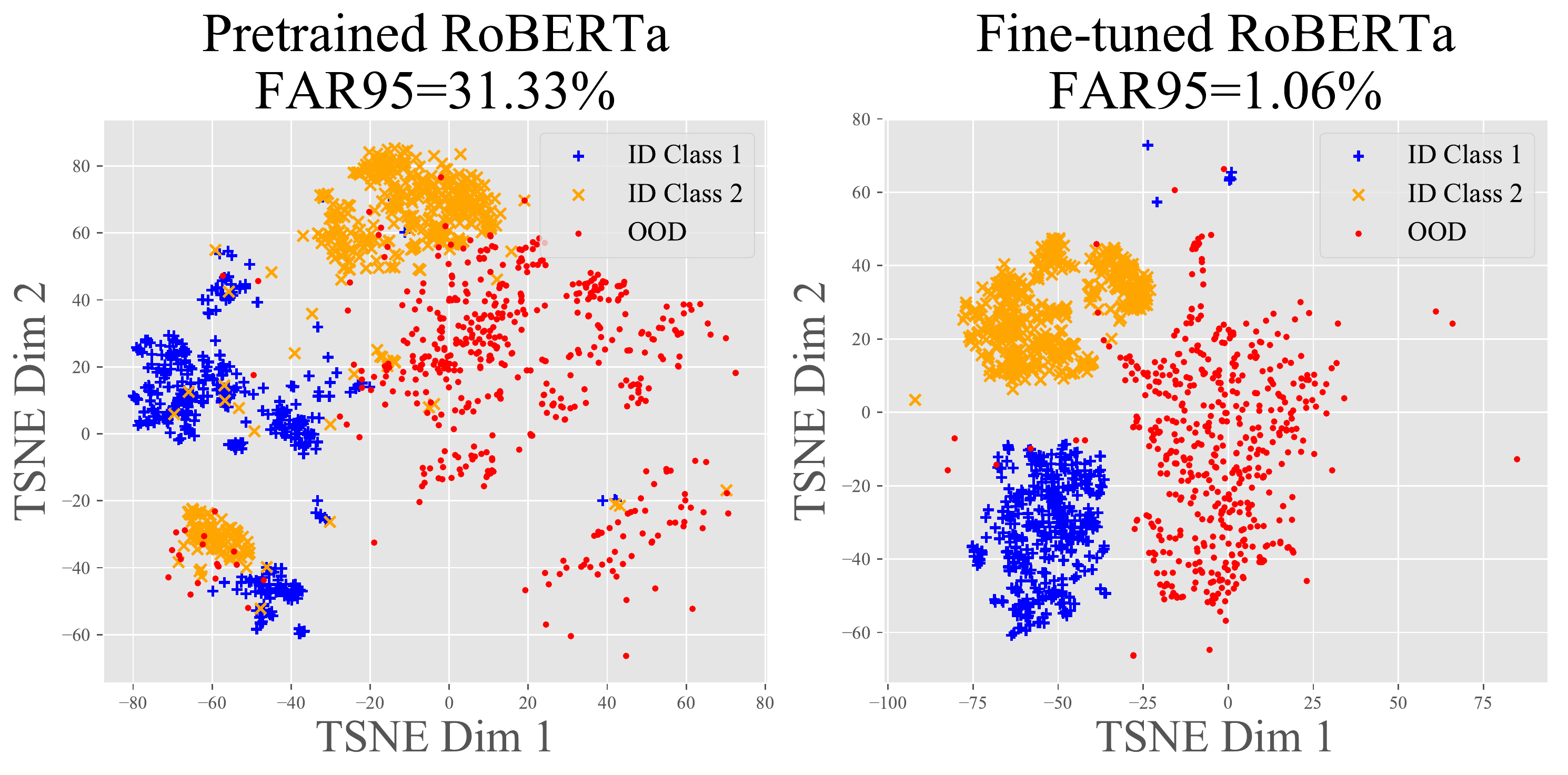}
         \label{fig1b}}
    \end{subfigure}
    
    \caption{T-SNE visualizations for the features derived from pre-trained and fine-tuned RoBERTa models and the corresponding FAR95 of the Mahalanobis detector.}
    \label{fig:tsne_features}
\end{figure}

To intuitively understand the effect of fine-tuning, we visualize the features using t-SNE \citep{van2008visualizing} in both settings. 
As plotted in Figure~\ref{fig:tsne_features}, before fine-tuning, the ID and OOD samples are sharply separated in the NSS setting, but show a significant overlap in the SS setting;
after fine-tuning, the ID samples are well clustered on class in both settings and unseen classes (OOD) in the SS setting are also pulled away from the ID data, but OOD samples in the NSS setting become almost indistinguishable from the ID data.

The observed benefits of fine-tuning in the SS setting match the observation in the near-OOD image detection~\citep{fort2021exploring}, suggesting that the fine-tuned task-specific representations are more suitable for detecting unseen classes belonging to the ID task.
Regarding the negative effect of fine-tuning in the NSS setting, we speculate that it can be explained in this way: task-agnostic features important for detecting non-semantic shifts are learned during pre-training but discarded in the fine-tuning stage, for which we will present empirical evidence in \S~\ref{subsec:3.5}.
%
To our knowledge, we are the first to study the impact of fine-tuning on the detection of different kinds of OOD texts and reveal the trade-off between the NSS setting (fine-tuning harms) and the SS setting (fine-tuning helps).

In addition, we notice that when the model is fine-tuned with margin-based contrastive auxiliary targets ($ \mathcal{L}_{\text{margin}}$) \citep{contrastive_nlp_ood}, the MD detector (MD$_{\text{ft}}$+$\mathcal{L}_{\text{margin}}$) substantially surpasses MD$_{\text{ft}}$ in the NSS setting with marginal sacrifice in the SS setting, thus it achieves the best performance on average.
However, it still falls far behind MD$_{\text{pre}}$ in the NSS setting.
\emph{As no single existing method behaves well in both settings, a general textual OOD detection method capable of detecting different kinds of OOD texts is yet to be developed, given the broad range of distribution shifts in realistic scenarios.}

\begin{figure}[t]
\centering
\includegraphics[width=0.48\textwidth]{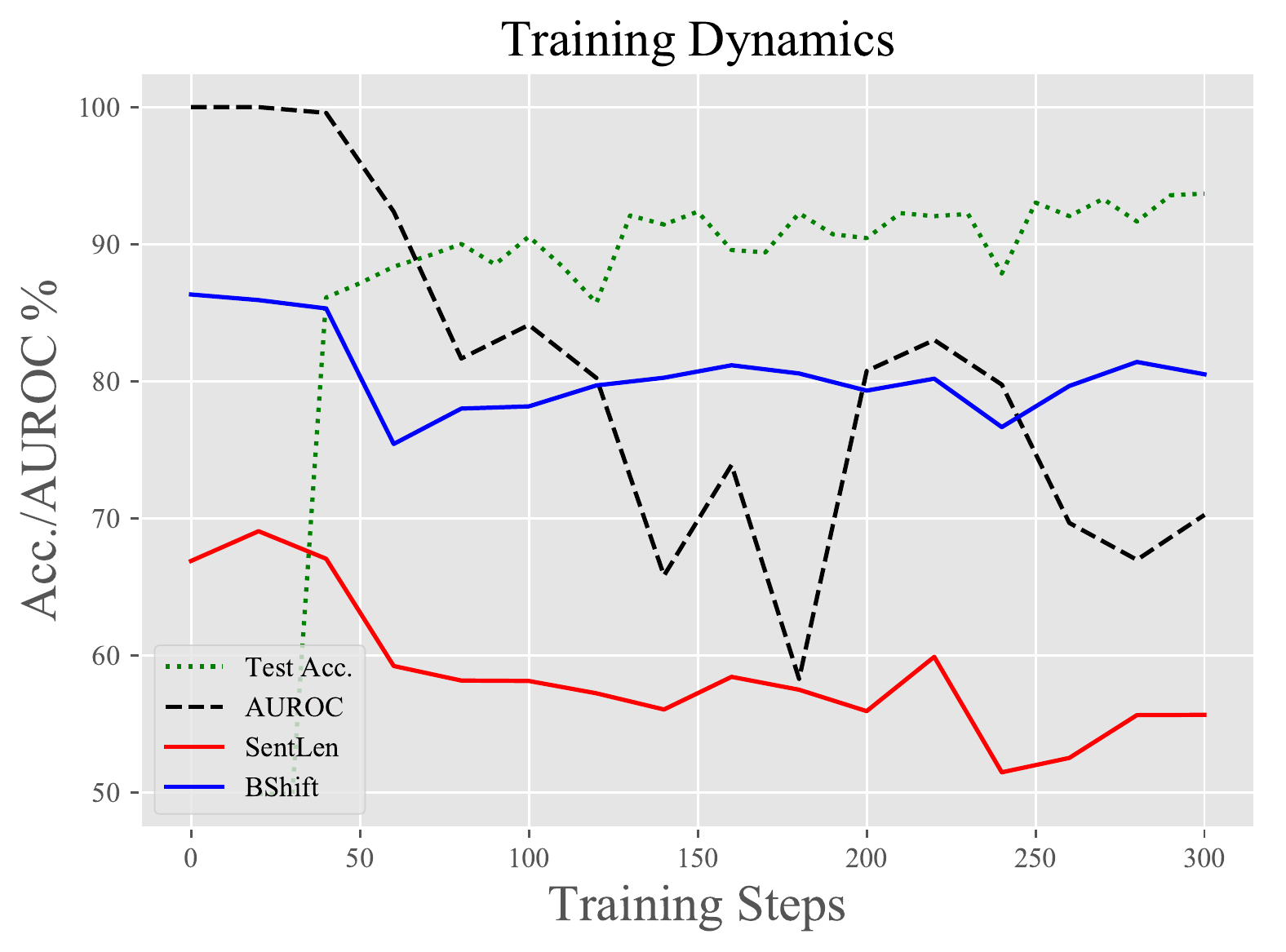}
\caption{The dynamics of test accuracy, probing accuracies of \textit{SentLen} and \textit{BShift}, and the performance (AUROC$\uparrow$) to detect IMDB as OOD in the fine-tuning process of the RoBERTa model on SST-2. We have seen similar trends on other datasets in the NSS setting.} 
\label{fig:probing}
\end{figure}

\subsection{Empirical Explanations} \label{subsec:3.5}

\paragraph{Probing Analysis.} 
From the view of the oracle, the OOD data in the NSS setting can be easily distinguished from the ID data by certain task-agnostic linguistic features. 
For example, the IMDB data are long movie reviews with an average length of 230 tokens, which can be well distinguished by length from the SST-2 reviews with an average length of 19 tokens.
Therefore, we speculate that the negative effect of fine-tuning arises from the deletion of general linguistic features during fine-tuning. 
To test the conjecture, we evaluate the sentence embeddings produced by the model checkpoints in the fine-tuning process on SST-2 on two classic probing tasks designed by \citet{conneau-kiela-2018-senteval}: \textit{SentLen} (sentence length) and \textit{BShift} (bigram shift).
 They are general linguistic features irrelevant to the class labels in downstream classification tasks, so the probing accuracies can be regarded as indicators of the preservation of task-agnostic features (see details in Appendix~\ref{app:probing_data}).
We show the tendency of the probing accuracies along with corresponding OOD detection performance (IMDB as OOD) and test accuracy in Figure~\ref{fig:probing}.
We find that as fine-tuning goes on, although the classification performance on ID test data shows an upward trend, the OOD detection performance (AUROC) gradually declines along with the probing accuracies.
The observed correlations between the OOD detection performance and the probing accuracies on \textit{SentLen} and \textit{BShift} empirically indicate that fine-tuning impairs NSS detection by distorting the task-agnostic features in pre-trained models.
\looseness=-1
\begin{table}[t]
\small
\centering
\begin{tabular}{@{}l|c|cc@{}}
\toprule
\textbf{Methods} & \textbf{Avg.} & \textbf{NSS} & \textbf{SS} \\ \midrule
MD$_{\text{pre}}$   &  50.29 & \textbf{29.90}   &  70.68  \\
MD$_{\text{ft}}$   &  57.42 & 80.95  & 33.89   \\
MD$_{\text{ft}}$ + RecAdam   &  55.76  & 78.39   & 33.13   \\
MD$_{\text{ft}}$ (head lr $\times$ 10)  & 53.90  & 76.18   & \textbf{31.63}   \\
MD$_{\text{ft}}$ + LP-FT  &  50.92   & 60.26  &  41.59  \\
                                     \bottomrule
\end{tabular}
\caption{The performance of MD$_{\text{ft}}$ coupled with regularization techniques on textual OOD detection. We report the FAR95$\downarrow$ values on average.}
\label{tab:finetuning_comparison}
\end{table}



\paragraph{Does Regularization Help?} 

As stated, fine-tuning may destruct the pre-trained features and thus harm NSS detection.
If the cause-effect holds, the negative effect can be alleviated by regularization techniques for preserving pre-trained features. 
To verify this deduction, we investigate three regularization approaches: (1) the RecAdam optimizer~\citep{RecAdam}; (2) a 10$\times$ larger learning rate for the head~\citep{prabhu2021sentry}; (3) the linear-probing then fine-tuning approach (LP-FT)~\citep{lpft}.
As shown in Table~\ref{tab:finetuning_comparison}, the regularization techniques applied to MD$_{\text{ft}}$ bring moderate improvements in the NSS setting, but they still fall far behind MD$_{\text{pre}}$ that exploits the original pre-trained features.
The results provide further empirical support for our reasoning about the demerits of fine-tuning.
Moreover, they suggest that effectively preserving pre-trained task-agnostic features suitable for NSS detection in fine-tuned PLMs is challenging.
A plausible solution is  to decouple task-agnostic and task-specific features in a single fine-tuned model, which we leave for future study.
Another possible solution is to directly leverage the pre-trained model, which we will introduce next.

\begin{table*}[t]
\centering
\resizebox{0.95\textwidth}{!}{
\begin{tabular}{l|c|cccc|cccc}
\toprule
\multirow{2.5}{*}{\textbf{Methods}}   & \multirow{2.5}{*}{\textbf{Avg.}}  & \multicolumn{4}{c}{\textbf{Non-Semantic Shift (NSS)}} & \multicolumn{4}{|c}{\textbf{Semantic Shift (SS)}}  \\    \cmidrule(lr){3-6} \cmidrule(lr){7-10}
 &  & \textbf{SST-2}    &   \textbf{IMDB}   & \textbf{Twitter}     & \textbf{Jigsaw}    & \textbf{NC}    &  \textbf{AGNews}    & \textbf{ROSTD}     &  \textbf{CLINC}     \\ \midrule
MSP \citep{maxprob}   & 72.92 &   90.50  & 79.20     & 89.76     &    96.29 & 72.16     &  85.06    &  51.24    & 19.08  \\
Scaling \citep{odin} & 71.62 &   90.50  & 79.20    &  89.76     & 96.29    &    68.94 & 80.19     &  52.15     & 15.90      \\
Energy \citep{liu2020Energy}  & 71.63 &  90.58   &  79.65     &  89.70   & 96.27    & 69.35     & 79.17     & 52.53     & 15.76     \\
D2U \citep{yilmazd2u} &  71.49 &  90.50   & 79.20    & 89.76     & 96.29    &     68.86 &  79.49    & 52.14    &   15.66  \\
PPL \citep{types} & 79.61  &  64.65   &  7.25   &  94.53    & 96.79   &  93.50    &  95.78   &    86.99 &  97.40    \\
LOF \citep{lin2019deep} & 58.82 &  93.39   & 24.28     & 88.86     & 92.90    & 70.83     & 81.00     & 4.69     &   14.58   \\ 
$\text{MD}_{\text{ft}}$ \citep{podolskiy2021revisiting} & 57.42 &  91.28   & 47.78     &88.49     &  96.25   & 58.16     & 64.06     &  1.06    & \textbf{12.26}     \\ 
$\text{MD}_{\text{ft}} $ + $ \mathcal{L}_{\text{scl}} $ \citep{contrastive_nlp_ood} & 60.36    &  88.05   &  63.11    & 84.53    & 94.57     &  67.00   &   69.26  & 3.42   & 12.90  \\
$\text{MD}_{\text{ft}} $ + $ \mathcal{L}_{\text{margin}} $  \citep{contrastive_nlp_ood} & 44.63  & 32.61    & 4.70      &72.51     &  96.90   &  59.31     & 67.48      & 11.07     & 12.40       \\
$\text{MD}_{\text{pre}}$ \citep{xu2021unsupervised} & 50.29  & 0.01    &    1.54 &  44.40    &\textbf{73.65} & 90.14     & 87.64 &  31.33   &  73.30   \\
$\text{KNN}_{\text{ft}}$ \citep{sun2022knn} & 56.91  & 87.87    &   56.81 & 83.67    & 95.60 & \textbf{56.51}     & \textbf{60.64} &  \textbf{0.71}   &  13.50  \\
$\text{KNN}_{\text{pre}}$ \citep{sun2022knn} & 46.29 &\textbf{0.00}    &    \textbf{1.48} & \textbf{33.11}    & 80.03 & 84.56     & 81.89 &  18.51  &  70.70   \\
GNOME (Ours)  &  \textbf{36.50}  &  0.04   & 8.24     &    53.36  &    85.88 &  64.81    & 63.25    & 1.47     & 14.94     \\\bottomrule
\end{tabular}}
\caption{OOD detection performance (FAR95$\downarrow$, lower is better) on the constructed suite of benchmarks. All values are percentages averaged over five different random seeds, and the best results are highlighted in \textbf{bold}. The second column gives the average performance on eight benchmarks.}
\label{tab:main_results}
\end{table*}




\section{GNOME for Textual OOD Detection}

In view of the observed trade-off, we are motivated to combine the strengths of task-agnostic and task-specific representations to obtain a confidence score capable of modeling both non-semantic shifts and semantic shifts. 
A straightforward way is to take the mean of MD$_{\text{pre}}$ and MD$_{\text{ft}}$ scores.\footnote{Our core idea is orthogonal to the distance-based scoring function, so we can also combine KNN$_{\text{pre}}$ and KNN$_{\text{ft}}$. We have tested in this way and got similar results to those obtained by combining MD$_{\text{pre}}$ and MD$_{\text{ft}}$.}
However, given that the norm of features can fluctuate and thus the distance scores are not comparable across different spaces, simple averaging may cause the integrated score to be skewed towards the side with the larger norm.
To alleviate the issue, we normalize MD$_{\text{pre}}$ and MD$_{\text{ft}}$ before aggregation:
\begin{equation}
    \begin{gathered}
    \operatorname{Norm}(\text{MD}_{\text{pre}}(\mathbf{x})) = \frac{\text{MD}_{\text{pre}}(\mathbf{x})-\mu_{\text{pre}}}{\sigma_{\text{pre}}}, \\
    \operatorname{Norm}(\text{MD}_{\text{ft}}(\mathbf{x})) = \frac{\text{MD}_{\text{ft}}(\mathbf{x})-\mu_{\text{ft}}}{\sigma_{\text{ft}}}, 
    \end{gathered}
\end{equation}
where $\mu$ and $\sigma$ are the mean and standard deviation of Mahalanobis distance scores, respectively, which can be estimated on ID validation samples. 
Then we obtain the integrated score GNOME (\textbf{G}e\textbf{N}eral textual \textbf{O}OD \textbf{M}easurement scor\textbf{E}):
\begin{equation}
\centering
\resizebox{0.42\textwidth}{!}{$
S_{\text{GNOME}}(\mathbf{x}) = -\operatorname{Agg}(\operatorname{Norm}(\text{MD}_{\text{pre}}(\mathbf{x})),\operatorname{Norm}(\text{MD}_{\text{ft}}(\mathbf{x})),$} 
\label{eq:gnome}
\end{equation} 
where $\operatorname{Agg}$ is the aggregation operator (such as the mean or max).
We use the mean operator for aggregation in our main experiments.
Note that We do not use a weighted average because it is not possible to tune the weights when OOD data is unknown, which follows the mainstream setting in OOD detection. 
If the user has prior knowledge about the type of OOD data, he/she can train the weights to aggregate the scores.

\section{Experiments}

\subsection{Experimental Setup}

\paragraph{Benchmarks} Besides the eight benchmarks introduced in \S~\ref{subsec:3.2} that are categorized as either NSS or SS, we also evaluate GNOME and baselines in the cross-task setting \citep{hendrycks2020pretrained,contrastive_nlp_ood} where both kinds of shifts happen simultaneously.
Following \citet{contrastive_nlp_ood}, we choose SST-2 \citep{sst2} and 20 Newsgroups \citep{20news} as ID data, and regard a series of datasets from different tasks as OOD data: TREC-10 \citep{li-roth-2002-learning}, WMT-16 \citep{wmt16}, Multi30k \citep{elliott2016multi30k}, RTE \citep{rte}, and SNLI \citep{snli}.  
Refer to Appendix~\ref{app:datasets} for more details.

\paragraph{Models and Metrics} We follow the same model configuration as that in \S~\ref{subsec:3.3} in main experiments and report FAR95 values in the main text (the trend of AUROC results in Appendix~\ref{app:results} is similar).

\begin{table}[t]
\centering
\small
\begin{tabular}{@{}l|c|cc@{}}
\toprule
\multirow{2.5}{*}{\textbf{Methods}} &  \multirow{2.5}{*}{\textbf{Avg.}} & \multicolumn{2}{c}{\textbf{ID Datasets}}    \\ \cmidrule(l){3-4}
 &  & \textbf{SST-2}      & \textbf{20 NG}      \\
\midrule
MSP  & 59.98 & 70.00 & 49.95  \\
Scaling & 50.68    &  70.00     & 31.36     \\
Energy & 52.31 & 72.43 & 32.31  \\
D2U  &    51.15   & 70.00 &  32.29       \\
LOF & 51.55 & 66.29 & 36.81  \\
$\text{MD}_{\text{ft}}$ & 32.29  & 48.82 & 15.75 \\
$\text{MD}_{\text{ft}} $ + $ \mathcal{L}_{\text{scl}} $  & 35.30 & 49.04 & 21.56  \\
$\text{MD}_{\text{ft}} $ + $ \mathcal{L}_{\text{margin}} $     & 23.97 & 29.43 & 18.51 \\
$\text{MD}_{\text{pre}}$ & 17.90 & 35.79 & \textbf{0.01}  \\
$\text{KNN}_{\text{ft}}$ & 43.58 & 63.73  & 23.42 \\
$\text{KNN}_{\text{pre}}$ & 20.79  & 41.57 & \textbf{0.01}  \\
GNOME (ours) & \textbf{13.02} & \textbf{26.02} & \textbf{0.01}   \\
\bottomrule
\end{tabular}
\caption{OOD detection performance in the cross-task setting. For each ID dataset, we report the macro average of FAR95$\downarrow$ on all corresponding OOD datasets, averaged over five random seeds.  }
\label{tab:main_results_cross_dataset}
\end{table}

\subsection{Results and Analysis}

\paragraph{GNOME works well in both SS and NSS settings and significantly surpasses baselines in terms of average performance.} As shown in Table~\ref{tab:main_results}, GNOME is competent in both settings (close to MD$_{\text{pre}}$ for NSS and MD$_{\text{ft}}$ for SS) and achieves the best performance on average. The average FAR95 is 36.50\%, 8.13\% lower than the previous SOTA MD$_{\text{ft}}$+$\mathcal{L}_{\text{margin}}$ requiring extra margin-based targets.

\paragraph{GNOME also achieves superior performance in the cross-task setting.} As results in Table~\ref{tab:main_results_cross_dataset}, GNOME outperforms all baseline methods (4.88\% FAR95 reduction on average) in the cross-task setting, demonstrating the power of integrating task-agnostic and task-specific representations when non-semantic shifts and semantic shifts happen simultaneously. Note that among existing methods, MD$_{\text{pre}}$ and KNN$_{\text{pre}}$ are the best on the 20 Newsgroups benchmark, suggesting that non-semantic shifts dominate there;  MD$_{\text{ft}}$+$\mathcal{L}_{\text{margin}}$ is the best on the SST-2 benchmark, indicating that both kinds of shifts matter there. 
Without any prior knowledge about the type of distribution shifts, GNOME yields the best performance on both benchmarks.

\subsection{Ablation Study} 
We examine the rationality of the key components of GNOME here. 
As shown in Table~\ref{table:ablation}, when the normalization operation is absent, the performance in the SS setting is slightly enhanced ($\sim$3\% FAR95 reduction), but the performance in the NSS and cross-task settings drops by around 7\% FAR95 points, which suggests that the normalization operation helps strike a balance between the two scenarios and thus achieve better performance on average.
These results also empirically verify that the mean operator is more suitable than the max operator for the score aggregation step in GNOME.
We have also tested other common normalization methods such as min-max and found that they underperform the standardization normalization employed in GNOME.\looseness=-1

Besides the score-level fusion in GNOME, we have also tested feature-level fusion (concatenating or averaging pre-trained and fine-tuning features), but they lead to a significant drop in the SS setting (+20\% FAR95) while only a slight improvement in the SS setting. 
Thus we argue that the score-level fusion by the mean operator is better.

\begin{table}[t]
\centering
\tiny
\resizebox{0.45\textwidth}{!}{
\begin{tabular}{@{}cc|c|ccc@{}}
\toprule
\textbf{Norm.} & \textbf{Agg.} &  \textbf{Avg.}  & \textbf{NSS} & \textbf{SS} & \textbf{CT}  \\ \midrule
\multirow{2}{*}{\Checkmark}      & mean     &     \textbf{28.67}          &      \textbf{36.88}        &      36.12     & \textbf{13.02}           \\
                       & max     & 30.08           &     38.33         &      37.90     & 14.01    \\ \midrule
\multirow{2}{*}{\XSolidBrush}      & mean       & 32.61         &   44.00           &     \textbf{33.32}               & 20.52   \\
                       & max        & 34.67       &   45.31           &     33.88       & 24.83    \\ \bottomrule 
\end{tabular}}
\caption{The performance (FAR95$\downarrow$) corresponding to different normalization choices and score aggregators in GNOME. CT denotes the cross-task setting.}
\label{table:ablation}
\end{table}
\begin{table}[t]
\centering
\small
\begin{tabular}{@{}lc|c|cc@{}}
\toprule
\textbf{Methods} & \textbf{\#Passes} & \textbf{Avg.}  &  \textbf{NSS} & \textbf{SS}  \\ \midrule
\textit{Single Pass} & & & & \\
MSP & 1 & 72.92 &  88.94 &  56.89  \\
MD$_{\text{ft}}$            &  1 &  57.42 &  80.95 & 33.89    \\
MD$_{\text{pre}}$            &  1 &  50.29 & \textbf{29.90} & 70.68  \\

\midrule
\textit{Model Ensemble}  &  &  & &  \\

MSP             & 2                 &   70.25 &     88.45  &  52.05          \\
MD$_{\text{ft}}$            &  2         & 56.09       &   81.35 & 30.82            \\
MSP             & 5  &    68.59 &  88.37   & 48.80                   \\
MD$_{\text{ft}}$    &   5  & 55.35      &  80.64   & \textbf{30.11}      \\

\midrule
\textit{Dropout Ensemble} &  &  & & \\
MC Dropout & 2  & 72.68 & 88.41 & 56.95   \\
MC Dropout & 5 &  70.75 & 85.88 & 55.62 \\

\midrule
GNOME           & 2           & \textbf{36.50}        &       36.88  & 36.10     \\ \bottomrule
\end{tabular}
\caption{Comparison with ensemble methods on the developed benchmark. We report FAR95 values averaged on the ID/OOD pairs in both SS and NSS settings.}
\label{tab:ensemble_comparsion}
\end{table}


\section{Further Discussion}

\subsection{Comparison with Ensemble Methods} \label{subsec:ensemble}

On the top of MD$_{\text{ft}}$ based on the fine-tuned PLM, GNOME is free of modification to the model architecture or training, and only requires an extra inference of the off-shelf PLM to obtain pre-trained features, thus being practical for real-world deployment.
For a strictly fair comparison under the same inference overhead constraint, we compare GNOME with previous ensemble methods, which can be divided into two groups: 
(1) \textit{Model ensemble} \citep{de}: summing confidence scores derived from models trained over different random seeds (we apply it to MSP and MD$_{\text{ft}}$);
(2) \textit{MC Dropout} \citep{gal2016mcdropout}: summing the probabilities output by multiple inferences with dropout on.

As the results shown in Table~\ref{tab:ensemble_comparsion}, previous ensemble methods that require 2$\times$ or 5$\times$ forward passes only slightly raise the performance compared with their single-pass counterparts, and fall far behind GNOME in terms of the average detection performance. 
These results also substantiate the power of integrating pre-trained and fine-tuned features.
We do not compare with the $k$-Folden method \citep{li2021kfolden} that needs $(C-1)$ sub-models ($C$ is the number of ID classes) because it does not apply to binary classification problems and is expensive for large-scale problems where $C$ is large.

\subsection{The Choice of Pre-Trained Features}

In the main experiments, we adopt the last-layer CLS embeddings as the pre-trained features for simplicity and fair comparison between MD$_{\text{pre}}$ and MD$_{\text{ft}}$.
As works on unsupervised textual OOD detection~\citep{xu2021unsupervised} and unsupervised sentence embedding~\citep{su2021whitening} show, pooling operations such as token-level and 
layer-level averaging produce better pre-trained features.
We then alternatively use \textit{last-avg} (the average of token embeddings in the last layer) and \textit{first-last-avg} (the average of token embeddings in the first and last layers) embeddings as pre-trained features in MD$_{\text{pre}}$ and GNOME.
As shown in Table~\ref{tab:pretrained_features}, when the \textit{last-cls} embeddings are replaced with the \textit{last-avg} or \textit{first-last-avg} embeddings, MD$_{\text{pre}}$ is moderately degraded in the NSS setting ($\sim$7\% FAR95 increase), but it is drastically improved in the SS setting ($\sim$14\% or $\sim$23\% FAR95 reduction). 
Notably, the trade-off before and after fine-tuning still holds when the \textit{last-avg} or \textit{first-last-avg} is used to get pre-trained features. 
However the pre-trained features are derived, GNOME consistently brings improvements to the average detection performance.

\begin{table}[t]
\small
\centering
\begin{tabular}{@{}cc|c|cc@{}}
\toprule
\textbf{\begin{tabular}[c]{@{}c@{}}Pre-trained \\ Features\end{tabular}} & \textbf{Methods} & \textbf{Avg.} & \textbf{NSS} & \textbf{SS} \\ \midrule
-   &  MD$_{\text{ft}}$ &   57.42    &  80.95 & 33.89    \\ \midrule
\multirow{2}{*}{\textit{last-cls}}    & MD$_{\text{pre}}$  & 50.29     & \textbf{29.90}          & 70.68 \\
& GNOME  & 36.50   &    36.88     & 36.10 \\ \midrule
\multirow{2}{*}{\textit{last-avg}}    & MD$_{\text{pre}}$  & 46.28  & 36.14         & 56.41 \\
& GNOME  & \textbf{35.88}  & 38.33  &  \textbf{33.43} \\ \midrule
\multirow{2}{*}{\textit{first-last-avg}}    & MD$_{\text{pre}}$  & 41.77 & 36.00         &  47.54 \\
& GNOME  & 35.89  &  37.93        &  33.85 \\  \bottomrule 
\end{tabular}
\caption{The OOD detection performance (FAR95$\downarrow$ in percentage) of different pre-trained features.}
\label{tab:pretrained_features}
\end{table}





\subsection{Generalization on Other PLMs}

To demonstrate the generality of GNOME, we also test on another two PLMs: BERT$_{\text{base-uncased}}$ \citep{devlin-etal-2019-bert} (110M parameters) and RoBERTa$_{\text{large}}$ \citep{roberta} (355M parameters).
As shown in Table~\ref{tab:backbone_ablation}, we observe that:
(1) The NSS-SS trade-off is prevalent on different PLMs and GNOME brings consistent gains over baselines in terms of average performance.
(2) RoBERTa$_{\text{base}}$, which uses more diverse pre-training data, beats BERT$_{\text{base-uncased}}$, suggesting that pre-training on diverse data boosts textual OOD detection; RoBERTa$_{\text{large}}$ underperforms RoBERta$_{\text{base}}$, indicating that larger models are not necessarily better at OOD detection.

\begin{table}[t]
\small
\centering
\begin{tabular}{@{}ll|c|cc@{}}
\toprule
\textbf{Backbone}                 & \textbf{Methods} & \textbf{Avg.} & \textbf{NSS} & \textbf{SS} \\ \midrule
\multirow{3}{*}{BERT$_{\text{base-uncased}}$}    &  MD$_{\text{pre}}$      &     67.81  &  60.61    &  75.01   \\
                         &  MD$_{\text{ft}}$      &  59.06    &  85.05   & 33.08    \\
                         &  GNOME      &  50.98    &    65.42 & 36.55   \\ \midrule
\multirow{3}{*}{RoBERTa$_{\text{base}}$} &  MD$_{\text{pre}}$      &  50.29    &    29.90 & 70.68    \\
                         &  MD$_{\text{ft}}$      &  57.42     &  80.95     &  33.89  \\
                         &  GNOME      & 36.50     & 36.88     & 36.10    \\ \midrule
\multirow{3}{*}{RoBERTa$_{\text{large}}$} & MD$_{\text{pre}}$       &  71.17    & 58.72    &  83.60  \\
                         & MD$_{\text{ft}}$        &   58.98   & 83.33     &     34.62 \\
                         &   GNOME     &  44.92      & 54.08     &  35.76   \\ \bottomrule
\end{tabular}
\caption{Textual OOD detection performance (FAR95$\downarrow$ values on average) with different pre-trained backbones.}
\label{tab:backbone_ablation}
\end{table}



\section{Conclusion}

Aware of the lack of a fair and comprehensive evaluation of current textual OOD detection methods, we take the first step to systematically assess them under different distribution shifts. 
Interestingly, we find that no single method works well in both the non-semantic shift setting and the semantic shift setting, and there exists a trade-off: fine-tuning pre-trained language models on in-distribution data benefits detecting semantic shifts but undermines detecting non-semantic shifts. 
After presenting empirical explanations for the trade-off from the perspective of feature distortion, we are then motivated to fully utilize both the pre-trained and fine-tuned features to obtain an efficient measurement score GNOME for better detecting diverse distribution shifts. 
Extensive experimental results demonstrate the efficacy and generality of GNOME.
Overall, GNOME is a first step in leveraging the intuition from our observations and analysis, and we hope that this work sheds light on the behavior of pre-trained language models upon detecting different kinds of distribution shifts and inspires new methods for general textual OOD detection.\looseness=-1






\section*{Limitations}

Although our approach GNOME  yields the best overall performance on the suite of benchmarks where either NSS or SS dominates and also performs best in the cross-task setting where both kinds of shifts take place, it slightly underperforms MD$_{\text{pre}}$ 
 and KNN$_{\text{pre}}$ in the NSS setting and marginally lags behind MD$_{\text{ft}}$ and KNN$_{\text{ft}}$ in the SS setting.
This is comprehensible because it is challenging for a single method to function perfectly for arbitrary OOD data without priors on the type of distribution shifts as analyzed in visual OOD detection works~\citep{ahmed2020semantic}.
Note again that we do not intend to present a perfect textual OOD detector capable of tackling all kinds of distribution shifts; instead, our core contributions are that we discover the trade-off between NSS and SS settings, present an empirical analysis to explain the phenomenon and provide insights to mitigate the trade-off for general textual OOD detection.\looseness=-1

\section*{Ethical Considerations}

We believe that our work leads to a better understanding of the behavior of pre-trained language models on OOD texts.
We also believe that the proposed method will facilitate the reliable deployment of NLU models since a model may face various types of OOD inputs in the wild and our method contributes to the detection performance on unknown OOD data in the average sense. 
All experiments in this work are conducted on open datasets and all pre-trained models that we investigate are publicly available.
We do not anticipate any negative social consequences to our work and we hope to continue to build on our method and develop more effective textual OOD detectors in the future. \looseness=-1

\section*{Acknowledgement}
We sincerely thank all the anonymous reviewers
for their valuable comments and advice.
This work is supported by Natural Science Foundation of China (NSFC) No. 62176002.
Xu Sun is the corresponding author of this paper.

\bibliography{anthology,custom}
\bibliographystyle{acl_natbib}

\appendix

\section{Dataset Introduction and Statistics} \label{app:datasets}

\subsection{The Constructed Suite of Benchmarks}

We show the included datasets in Table~\ref{tab:dataset_stats} and give an introduction to them as follows.

For the NSS setting, we consider two tasks: sentiment analysis and toxic detection.
For sentiment analysis, we choose SST-2 \citep{sst2} and IMDB \citep{imdb}.
SST-2 contains short movie reviews by the audience, while IMDB contains longer and more professional movie reviews. Therefore, the two datasets can regard each other as OOD data representing a non-semantic shift.
For toxic detection, we choose Twitter \citep{twitter} and  the Jigsaw dataset from a Kaggle challenge.\footnote{Available at this \href{https://www.kaggle.com/competitions/jigsaw-toxic-comment-classification-challenge/data}{link}.}   
The Twitter dataset consists of short comments on Tweet, while the Jigsaw dataset consists of longer Wikipedia comments, so they can regard each other as OOD data of the NSS type.

For the SS setting, we consider two tasks where newly emerging classes are common:  news topic categorization and dialogue intent classification.
For news topic categorization, we choose the AGNews \citep{agnews} and News Category datasets \citep{misra2018news} to construct ID/OOD pairs. Specifically, we use 4 classes from AGNews and 5 classes from News Category  
as ID data and use the remaining classes from the original datasets as OOD data. 
For dialogue intent classification, we use the ROSTD \citep{gangal2020likelihood} and CLINC \citep{clinc} datasets as ID data and use the annotated unknown intents from the original datasets as OOD data.

\begin{table}[t]
\centering
\resizebox{0.48\textwidth}{!}{
\begin{tabular}{@{}llllll@{}}
\toprule
\textbf{Dataset} & \textbf{\# Classes} & \textbf{\# Train} & \textbf{\# Dev} & \textbf{\# Test} & \textbf{L} \\ \midrule
SST-2            &     2             &                6,920      &        872         &    1,821              &        19    \\
IMDB             &        2          &               23,000       &        2,000         &  25,000                &      230      \\
Twitter       &     2              &     69,632           &    7,737      & 8,597               &   17        \\
Jigsaw          &      2              & 143,614              &  15,957             &       63,978 &     68    \\ 
AGNews  & 4  & 115,778  & 3,994 & 3,993  &  23   \\  
NC & 5 &   68,859     & 8,617 & 8,684 & 30 \\
    ROSTD    &     12 &  30,521 &    4,181 & 8,621 & 7  \\  
CLINC & 150 & 15,000 & 3,000 & 4,500 & 8 \\
AGNews$_{\text{OOD}}$ &  - & - & - & 3,600 &  21 \\
NC$_{\text{OOD}}$ &  - & - & - & 11,402 &  29 \\
ROSTD$_{\text{OOD}}$  &  - & - & - & 3,090 &  7 \\
CLINC$_{\text{OOD}}$  &  - & - & - & 1,000 &  9 \\
\bottomrule
\end{tabular}}
\caption{Statistics of the datasets used for the constructed suite of benchmarks. \textbf{L} denotes the average length of each sample.}
\label{tab:dataset_stats}
\end{table}

\subsection{Cross-Task Benchmarks}

For the cross-task setting, we follow \citet{contrastive_nlp_ood} to use SST-2 and 20 Newsgroups (20 NG) \citep{20news} as ID data. 
20 NG is a news categorization dataset containing 10,182 training samples, 1,132 validation samples, and 7,532 test samples. The average sample length in 20 NG is 289.
Naturally, SST-2 and 20 NG can regard each other as OOD data.
Besides, we use five additional datasets from different datasets as OOD test data for each ID dataset: TREC-10 \citep{li-roth-2002-learning}, WMT-16 \citep{wmt16}, Multi30k \citep{elliott2016multi30k}, RTE \citep{rte}, and SNLI \citep{snli}.  
TREC-10 is a question classification dataset; Multi30k \cite{elliott2016multi30k} and WMT16 \cite{wmt16} are parts of the English side data of English-German machine translation datasets; RTE \cite{rte} and SNLI \cite{snli} are the concatenations of the precise and respective hypotheses from NLI datasets.
The statistics of the OOD datasets are listed in Table~\ref{tab:ood_stats}.

\begin{table}[t]
\small          
\centering
\begin{tabular}{@{}lll@{}}
\toprule
\textbf{Dataset} & \textbf{\# Test} & \textbf{L} \\ \midrule
TREC-10         &   500   & 10  \\
Multi30k         &     1,014                  & 13 \\
WMT16            &   2,000                & 22           \\
RTE              &      3,000            & 48            \\
SNLI             &      2,000            &  21          \\ \bottomrule
\end{tabular}
\caption{Statistics of OOD datasets in the cross-task setting. \textbf{L} denotes the average length of each sample.}
\label{tab:ood_stats}
\end{table}

\subsection{Probing Benchmarks}  \label{app:probing_data}
To probe the linguistic information contained in pre-trained and fine-tuned features, we use two probing tasks designed by \citet{DBLP:conf/acl/BaroniBLKC18}.
Each probing dataset contains 100k training samples, 10k validation samples, and 10k test samples. 
We use the SentEval toolkit~\citep{conneau-kiela-2018-senteval} along with the recommended hyperparameter space to search for the best probing classifier according to the validation accuracy and report test accuracies.

\section{Performance on In-Distribution Data} \label{app:id_acc}

\begin{table}[t]
\centering
\small
\begin{tabular}{@{}lccc@{}}
\toprule
\textbf{Dataset / Loss} &  $\mathcal{L}_{\text{ce}}$    & $\mathcal{L}_{\text{ce}}$ + $\mathcal{L}_{\text{scl}}$   &  $\mathcal{L}_{\text{ce}}$ + $ \mathcal{L}_{\text{margin}}$ \\ \midrule
SST-2                  &    93.96         & 94.23              &    93.69             \\
IMDB                   &    94.56         & 94.53             &     94.21            \\
Twitter                &    93.67         &    93.64          &     93.81            \\
Jigsaw                 &    81.82         &   82.08           &     82.43            \\
NC          &       95.39      &       95.21       &         95.43        \\
AGNews                 &   91.28          &   91.03           &    91.18             \\
ROSTD                  &      99.23       &      99.21        &  99.26               \\
CLINC                  &      96.21       &    96.16          &  96.08                \\ 
20 NG               &    84.52 & 84.65 & 84.53 \\
\bottomrule
\end{tabular}
\caption{Accuracies / F1 scores on the test set of in-distribution data (averaged over five random seeds).  We report F1 scores for Twitter and Jigsaw toxic detection and accuracies for other tasks. }
\label{tab:test_acc}
\end{table}

We fine-tune the RoBERTa$_{\text{base}}$ model on the ID training data to build text classifiers in our main experiments.
The model is optimized with the Adam \citep{adam} optimizer using a learning rate of 2e-5.
 We use a batch size of 16 and fine-tune the model for 5 epochs.
We evaluate the model on the ID validation set after every epoch and choose the best checkpoint as the final model. 
The setting is the same for other pre-trained Transformers studied in the paper (BERT$_{\text{base-uncased}}$ and RoBERTa$_{\text{large}}$).
The performance of fine-tuned RoBERTa$_{\text{base}}$ models is given in Table~\ref{tab:test_acc}, where  $\mathcal{L}_{\text{ce}}$ denotes the vanilla cross-entropy loss, $\mathcal{L}_{\text{scl}}$ denotes the supervised contrastive loss \citep{khosla2020scl}, and $\mathcal{L}_{\text{margin}}$ denotes the margin-based contrastive loss~\citep{contrastive_nlp_ood}.
We report the F1 scores on the test set for toxic detection on Twitter and Jigsaw and test accuracies for other tasks.

\section{Details of OOD Detection Baselines} \label{app:details}


\subsection{Confidence-Based Baselines}

\paragraph{Notations} In a classification problem with $C$ classes, assume the input is $\mathbf{x}$, we denote $f_i(\mathbf{x})$ is the output logit of class $i$, and the predicted softmax probability of class $i$ is defined as:
\begin{equation}
    p_i(\mathbf{x}) =\max _{i} \frac{\exp \left(f_{i}(\boldsymbol{x}) \right)}{\sum_{j=1}^{C} \exp \left(f_{j}(\boldsymbol{x}) \right)}.
\end{equation}
Confidence-based methods obtain the OOD score based the output logits and softmax probabilities.

\paragraph{MSP} \citet{maxprob} propose the maximum softmax probabilty (MSP) baseline, in which the confidence score is defined the predicted maximum softmax probability among $C$ classes:
\begin{equation}
    S(\mathbf{x}) = \max_{i} p_i(\mathbf{x}). 
\end{equation}

\paragraph{Scaling} In the ODIN paper \citep{odin}, tempature scaling is applied to the scoring function:
\begin{equation}
    S(\mathbf{x}) = \max _{i} \frac{\exp \left(f_{i}(\boldsymbol{x})/T \right)}{\sum_{j=1}^{C} \exp \left(f_{j}(\boldsymbol{x})/T \right)},
\end{equation}
where $T$ is the temperature term.
Following \citet{godin}, we fix $T=1000$ in our experiments.
Note that ODIN \citep{odin} also propose an input pre-processing step adding adversarial perturbation to the input image, while we do not use it because it is not directly applicable for discrete inputs in NLP.

\paragraph{Energy Score}
\citep{liu2020Energy} propose to use the free energy function for OOD detection, which is formulated as follows:
\begin{equation}
    \begin{gathered}
E(\mathbf{x}) = \sum_{i=1}^{C} e^{f_i(\mathbf{x})}, \\
S(\mathbf{x}) = -E(\mathbf{x}).
    \end{gathered}
\end{equation}

\paragraph{D2U} \citet{yilmazd2u} propose to improve out-of-scope detection by exploiting the shape of the entire output distribution.
Specifically, the distance of the output distribition $P(\mathbf{x})=\left(p_1(\mathbf{x}),\ldots,p_c(\mathbf{x})\right)$ to the uniform distribution $U$ as the OOD score:
\begin{equation}
    S(\mathbf{x}) = \operatorname{dst}\left(P\left(x\right),U\right),
\end{equation}
where $\operatorname{dst}$ is the distance function. 
We use the KL divergence as the distance function as recommended in \citet{yilmazd2u} in our experiments.
Note that \citet{yilmazd2u} also propose to use D2U for loss calculation when out-of-scope training data is available, while we do not use it in the training because we follow the mainstream setting in OOD detection works where OOD data is not available for training.

\subsection{Density-Based Baselines}

\paragraph{PPL}  
\citet{types} propose to use the token perplexity (PPL) score derived from the GPT-2 language model \citep{gpt2} as the OOD score.
Following the implementation of \citet{types},\footnote{Available at this Github \href{https://github.com/uditarora/ood-text-emnlp}{repository}.} we fine-tune the GPT-2$_{\text{small}}$ model (117M parameters, similar to RoBERRa$_{\text{base}}$ in size) for language modeling on the ID training data and use the inverse of the PPL score as the OOD score. 
Formally, for an input text sequence $\mathbf{x} = \left\{x_1,\ldots,x_t\right\}$, 
\begin{equation}
    \begin{gathered}
        \operatorname{PPL}(\mathbf{x}) = \exp \left\{-\frac{1}{t} \sum_{i}^{t} \log p_{\theta}\left(x_{i} \mid x_{<i}\right)\right\}, \\
        S(\mathbf{x}) = 1/\operatorname{PPL}(\mathbf{x}),
    \end{gathered}
\end{equation}
where $t$ is the number of tokens in $\mathbf{x}$.

\subsection{Distance-Based Baselines}

\paragraph{Local Outlier Factor (LOF)} 
\citet{lin2019deep} propose to identify unknown user intents by feeding feature vectors derived from LSTM models to the density-based novelty detection algorithm, local outlier factor (LOF) \citep{breunig2000lof}. 
In our implementation, we use the last-layer CLS vector embeddings by the fine-tuned RoBERTa models as the input and train a LOF model following the implementation details of \citet{lin2019deep} on the ID training set.
Finally, we use the local density output as $S(\mathbf{x})$. 

\paragraph{Mahalanobis Distance Detector} The Mahalanobis distance detector (MD) \citep{maha} is a classical distance-based OOD detection method that exploits the sample distance to the nearest ID class in the embedding space to obtain the OOD score. 
Formally, for a given feature extractor $\psi$, the Mahalanobis distance score is defined as: 
\begin{equation}
\resizebox{0.40\textwidth}{!}{$
\centering
S(\mathbf{x})=-\min_{c \in \Upsilon}  \left(\psi(\mathbf{x})-\mu_{c}\right)^{T} \Sigma^{-1}\left(\psi(\mathbf{x})-\mu_{c}\right),$}
\end{equation}
where $\Upsilon = \left\{1,2,\ldots,C\right\}$ is the label space containing $C$ classes in the ID task, $\psi(\mathbf{x})$ is the embedding vector of the input $\mathbf{x}$, $\mu_{c}$ is the class centroid for a class $c$, and $\Sigma$ is the covariance matrix. 
The estimations of $\mu_{c}$ and $\Sigma$ are defined as:
\begin{equation}
\resizebox{0.40\textwidth}{!}{$
\begin{gathered}
\mu_{c}=\frac{1}{N_{c}} \sum_{\mathbf{x} \in \mathcal{D}_{i n}^{c}} \psi(\mathbf{x}), \\
\Sigma=\frac{1}{N} \sum_{c \in \Upsilon} \sum_{\mathbf{x} \in \mathcal{D}_{i n}^{c}}\left(\psi(\mathbf{x})-\mu_{c}\right)\left(\psi(\mathbf{x})-\mu_{c}\right)^{T},
\end{gathered}$}
\end{equation}
where $\mathcal{D}_{\text{in}}^{c}=\left\{\mathbf{x} \mid(\mathbf{x}, y) \in \mathcal{D}_{\text{in}}, y=c\right\}$ denotes the training samples belonging to the class $c$, $N$ is the size of the training set, and $N_{c}$ is the number of training instances belonging to the class $c$.
As for textual OOD detection based on pre-trained language models, when the feature extractor $\psi$ is the off-the-shelf pre-trained model, i.e. detecting anomalies in the pre-trained feature space \citep{xu2021unsupervised}, it is called MD$_{\text{pre}}$ in our paper; when $\psi$ is the fine-tuned model, i.e. detecting anomalies in the fine-tuned feature space \citep{podolskiy2021revisiting}, it is called MD$_{\text{ft}}$ in our paper.

\paragraph{Contrastive Fine-Tuning Targets Coupled with the MD Detector}

\citet{contrastive_nlp_ood} propose to use two forms of contrastive losses to boost textual OOD detection, i.e., the supervised contrastive loss ($\mathcal{L}_{\text {scl}}$) and the margin-based contrastive loss ($\mathcal{L}_{\text {margin}}$). 
For a classification task containing $C$ classes, given a batch of training examples $\left\{x_i,y_i\right\}_{i=1}^M$, where $x_i$ is the input and $y_i$ is the label, the supervised contrastive loss term $\mathcal{L}_{\text {scl}}$ and the final optimization target $\mathcal{L}$ can be formulated as:
\begin{equation}
\resizebox{0.40\textwidth}{!}{$
\begin{gathered}
  \mathcal{L}_{\text{scl}} =\sum_{i=1}^{M} \frac{-1}{M|P(i)|} \sum_{p \in P(i)} \log \frac{e^{\boldsymbol{z}_{i}^{\top} \boldsymbol{z}_{p} / \tau}}{\sum_{a \in A(i)} e^{\boldsymbol{z}_{i}^{\top} \boldsymbol{z}_{a} / \tau}}, 
\\
  \mathcal{L} = \mathcal{L}_{\text{ce}} +\mathcal{L}_{\text{scl}},
\end{gathered}$}
\end{equation}
where $A(i) = \{1, ...,M\} \setminus \{i\}$ is the set of all anchor samples, $P(i) = \{ p \in A(i): y_i = y_p \} $ is the set of anchor samples from the same class  as $i$, $\tau$ is a temperature hyper-parameter, $\boldsymbol{z}$ is the L2-normalized CLS embedding before the softmax layer, $\mathcal{L}_{\text {ce}}$ is the cross-entropy loss, and $\lambda$ is a positive coefficient.
Following the implementation of \citet{contrastive_nlp_ood},\footnote{Available at this Github \href{https://github.com/wzhouad/Contra-OOD}{repository}} we use $\tau=0.3$ and $\lambda=2$.

The margin-based loss term $\mathcal{L}_{\text {margin }}$ and the final optimization target $\mathcal{L}$ is formulated as:
\begin{equation}
\resizebox{0.40\textwidth}{!}{$
\begin{gathered}
\mathcal{L}_{\text {pos }}=\sum_{i=1}^{M} \frac{1}{|P(i)|} \sum_{p \in P(i)}\left\|\boldsymbol{h}_{i}-\boldsymbol{h}_{p}\right\|^{2}, \\
\mathcal{L}_{\text {neg }}=\sum_{i=1}^{M} \frac{1}{|N(i)|} \sum_{n \in N(i)}\left(\xi-\left\|\boldsymbol{h}_{i}-\boldsymbol{h}_{n}\right\|^{2}\right)_{+}, \\
\mathcal{L}_{\text {margin }}=\frac{1}{d M}\left(\mathcal{L}_{\text {pos }}+\mathcal{L}_{\text {neg }}\right), \\
\xi=\max _{i=1}^{M} \max _{p \in P(i)}\left\|\boldsymbol{h}_{i}-\boldsymbol{h}_{p}\right\|^{2}, \\
\mathcal{L} = \mathcal{L}_{\text{ce}} + \lambda \mathcal{L}_{\text{margin}},
\end{gathered}$}
\end{equation}
where $N(i)=\left\{n \in A(i): y_{i} \neq y_{n}\right\}$ is the set of anchor samples from other classes than $y_{i}$, $\boldsymbol{h} \in \mathbb{R}^d$ is the unnormalized CLS embedding before the classification head, $\xi$ is the margin, $d$ is the number of dimensions of $\boldsymbol{h}$, and $\lambda$ is a positive coefficient. 
We use $\lambda=2$ following \citet{contrastive_nlp_ood}. 

Except for the optimization target, we use the same hyper-parameters for the two tuning methods as vanilla tuning.

\paragraph{Nearst-Neighbor-Based Detector}
\citet{sun2022knn} explore the efficacy of non-parametric nearest-neighbor distance for OOD detection and show its advantages over the Mahalanobis distance detector on visual OOD detection benchmarks.
Specifically, it takes the minus of the average distance from the test sample to the $k$-nearest training samples in the normalized feature space. 
We reproduce two variants, i.e., $\text{KNN}_{\text{pre}}$ using the pre-trained features and $\text{KNN}_{\text{ft}}$ using the fine-tuned features. We set the neighborhood size $k=10$ in our experiments.

\begin{table*}[t]
\centering
\resizebox{0.95\textwidth}{!}{
\begin{tabular}{l|c|cccc|cccc}
\toprule
\multirow{2.5}{*}{\textbf{Methods}}   & \multirow{2.5}{*}{\textbf{Avg.}}  & \multicolumn{4}{c}{\textbf{Non-Semantic Shift (NSS)}} & \multicolumn{4}{|c}{\textbf{Semantic Shift (SS)}}  \\    \cmidrule(lr){3-6} \cmidrule(lr){7-10}
 &  & \textbf{SST-2}    &   \textbf{IMDB}   & \textbf{Twitter}     & \textbf{Jigsaw}    & \textbf{NC}    &  \textbf{AGNews}    & \textbf{ROSTD}     &  \textbf{CLINC}     \\ \midrule
MSP \citep{maxprob} & 71.62 &  67.92  &  74.09   & 48.75    &  71.13  &   75.12  & 64.84    &  75.42   & 95.72   \\
Scaling \citep{odin}     & 71.96   & 67.92    & 74.09     & 48.76     & 71.03   & 74.60    &  67.35    & 75.71     & 96.20 \\
Energy \citep{liu2020Energy} & 71.63 & 69.73    & 72.84     & 47.56     & 69.49    & 74.19     & 67.55     & 76.52     &    96.18   \\
D2U \citep{yilmazd2u} & 71.99  &  67.92   & 74.09     & 48.75     & 71.13    & 74.62     & 67.46     & 75.72     & 96.26      \\
PPL \citep{types}  &  67.65  &   79.65  & 98.51     & 34.61    &  \textbf{84.36}   &  57.91    & 50.67    & 85.05     & 50.47      \\
LOF \citep{lin2019deep} & 76.42 &  53.39   & 94.87     & 62.14     & 57.88    & 78.39     & 70.07     & 97.49     & 97.17     \\
$\text{MD}_{\text{ft}}$ \citep{podolskiy2021revisiting} & 80.39  &  69.86   & 90.87     &    65.83 & 62.42    & 3.41     & 73.51     & 99.66    & \textbf{  97.57  }   \\ 
$\text{MD}_{\text{ft}} $ + $ \mathcal{L}_{\text{scl}} $  \citep{contrastive_nlp_ood} &  82.22   &   83.12  &  88.28    & 71.13    & 64.32     & 81.68     &  72.74    & 99.09    & 97.39  \\
$\text{MD}_{\text{ft}} $ + $ \mathcal{L}_{\text{margin}} $   \citep{contrastive_nlp_ood} & 86.50    &  93.84   &  \textbf{98.99}    & 83.04    & 65.94     &  82.91    & 72.00     & 97.68    & 97.56  \\
$\text{MD}_{\text{pre}}$ \citep{xu2021unsupervised}  &  83.76  & \textbf{ 99.99 }   & 98.75  & 90.59    & 83.84   & 57.45    & 61.53    & 95.22     &   82.69    \\
$\text{KNN}_{\text{ft}}$ \citep{sun2022knn} & 81.02 &  72.00   &  86.07    &    74.90 & 57.33 & \textbf{84.82}    & \textbf{75.85}   &  \textbf{99.67}   & 97.53   \\ 
$\text{KNN}_{\text{pre}}$ \citep{sun2022knn} & 85.69  & \textbf{99.99}    &   98.31   &  \textbf{92.80}   & 79.53    &   65.90   & 69.28   & 96.89   & 82.81    \\ 
GNOME (Ours)  &  \textbf{89.34} &   99.98  & 98.25     & 89.64     & 81.10    & 75.50    & 73.77    & 99.63     & 96.84      \\\bottomrule
\end{tabular}}
\caption{OOD detection performance (AUROC$\uparrow$, higher is better) on the developed suites of benchmarks. All values are percentages averaged over five different random seeds, and the best results are highlighted in \textbf{bold}.  The last column gives the average performance on eight datasets.}
\label{tab:main_results_auroc}
\end{table*}



\section{Software and Hardware Requirements} 

We implement our code based on the PyTorch \citep{torch} and HuggingFace Transformers \citep{wolf2020transformers} Python libraries. 
All experiments (training and inference) in this paper can be conducted on a single NVIDIA TITAN RTX GPU (24 GB memory), except that the fine-tuning of the RoBERTa$_{\text{large}}$ model needs 4 TITAN RTX GPUs.

\section{Additional Experimental Results} \label{app:results}

We display the AUROC results of GNOME and the baselines on the constructed suite of benchmarks in Table~\ref{tab:main_results_auroc}.
The overall trend is consistent with that of the FAR95 results reported in Table~\ref{tab:main_results} in the main text.

\end{document}